\pdfoutput=1

\documentclass[11pt]{article}

\usepackage[]{ACL2023}

\usepackage{times}
\usepackage{latexsym}
\usepackage{multirow}
\usepackage{multicol}
\usepackage{booktabs}
\usepackage{graphicx}
\usepackage{amssymb}
\usepackage{makecell}
\usepackage{amsmath}
\usepackage[T1]{fontenc}

\usepackage[utf8]{inputenc}

\usepackage{microtype}

\usepackage{inconsolata}

\title{RFiD: Towards Rational Fusion-in-Decoder for \\Open-Domain Question Answering}

\author{
Cunxiang Wang\textsuperscript{$\clubsuit$}, 
Haofei Yu\textsuperscript{$\heartsuit$\thanks{\ \ Co-first Author}}, 
Yue Zhang\textsuperscript{$\clubsuit$\thanks{\ \ The correponding author.}}
\\
\textsuperscript{$\clubsuit$}School of Engineering, Westlake University, China\\
\textsuperscript{$\heartsuit$}Language Technologies Institute, Carnegie Mellon University, USA\\
  {\{wangcunxiang, zhangyue\}@westlake.edu.cn; haofeiy@cs.cmu.edu}
  }
\begin{document}

\maketitle
\begin{abstract}
Open-Domain Question Answering (ODQA) systems necessitate a reader model capable of generating answers by simultaneously referring to multiple passages. Although representative models like Fusion-in-Decoder (FiD) have been proposed to address this challenge, these systems can inadvertently rely on spurious features instead of genuine causal relationships between the question and the passages to generate answers. To counter this problem, we introduce the Rational Fusion-in-Decoder (RFiD) model. Our model leverages the encoders of FiD to differentiate between causal relationships and spurious features, subsequently guiding the decoder to generate answers informed by this discernment. Experimental results on two ODQA datasets, Natural Questions (NQ) and TriviaQA (TQ), demonstrate that our model surpasses previous methods, achieving improvements of up to 1.5 and 0.7 in Exact Match scores on NQ, and exhibits an enhanced ability to identify causal relationships.\footnote{Our code and data are available at https://github.com/wangcunxiang/RFiD}
\end{abstract}

\section{Introduction}

\begin{figure}[t]
  \center{
  \includegraphics
  [width=18cm]  
  {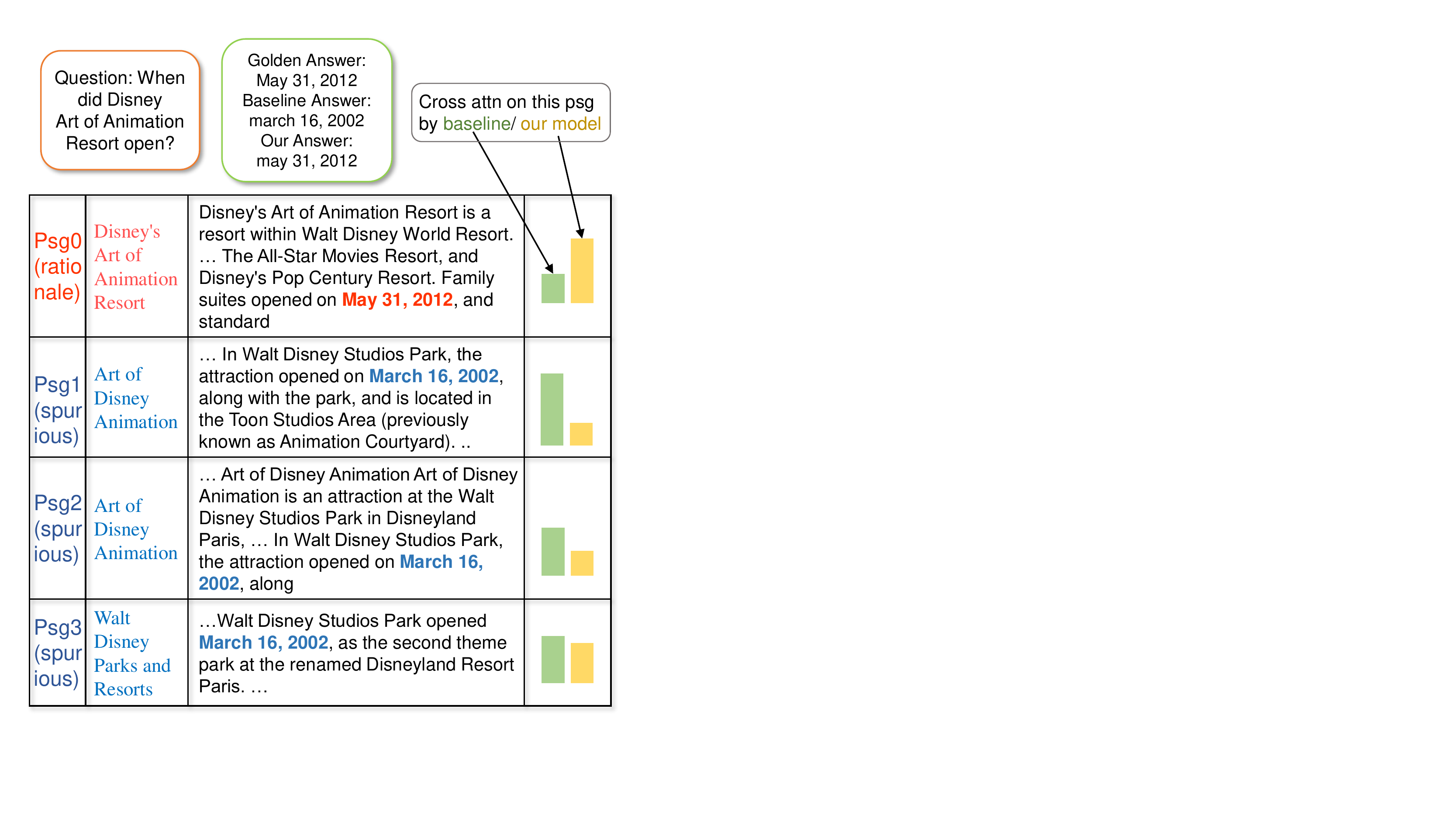}}
  \vspace{-18mm}
  \caption{
  An example from our experiments. The question has only one relevant passage (red Psg0), while the remaining blue ones represent three passages that contain the wrong answer generated by the baseline model. }
  \label{main_example}
  \vspace{-3mm}
\end{figure}

Open-domain Question Answering (ODQA) has garnered significant attention \citep{DrQA, NQ, TQ}, leading to the development of various systems designed to retrieve relevant passages \citep{DPR, SEAL} from large databases and generate corresponding answers \citep{FiD, rag}. We utilize the Fusion-in-Decoder (FiD) \citep{FiD} model as our baseline model, a sequence-to-sequence paradigm based on the T5 model \citep{t5}. Given a question, the FiD model encodes $K$ retrieved passages using $K$ respective T5 encoders, concatenates these $K$ encoder hidden states, and then feeds the result into a T5 decoder to generate the answer.

FiD treats all passages equally in its encoders, relying exclusively on the cross-attention mechanism to establish correlations between the decoder and encoders. However, the cross-attention mechanism lacks an explicit mechanism for distinguishing differences among passages, which can result in the detection of spurious patterns \citep{fooling, Jo2017MeasuringTT}. Consequently, it becomes challenging for the model to identify crucial passages. An example of such spurious patterns observed in our experiment is depicted in Figure~\ref{main_example}, where the model confuses "Disney Art of Animation Resort" with "Art of Disney Animation" due to the prevalence of passages about the latter, resulting in an incorrect answer.

To address this issue, we propose a conceptually straightforward strategy by introducing a rationalization process to explicitly determine which retrieved passages contain the answer before conducting answer generation. This process assigns different embeddings to rationale passages and irrelevant passages. These embeddings then guide the cross-attention during the answer generation phase. We dub this new model the \textbf{R}ational \textbf{F}usion-\textbf{i}n-\textbf{D}ecoder (RFiD).

We evaluate the effectiveness of our proposed RFiD model through experiments on the Natural Questions (NQ) \citep{NQ} and TriviaQA (TQ) \citep{TQ} datasets. Our results demonstrate that our methods can help models overcome spurious patterns and enhance their reasoning abilities, leading to an improvement of up to 1.5/0.7 Exact Match points on the NQ/TQ datasets respectively. Further analysis reveals that our methods effectively direct models to focus more on correct causal features and less on spurious features. For instance, as seen in the rightmost column of Figure~\ref{main_example}, our model has increased its attention on the relevant passage.

To the best of our knowledge, we are the first to incorporate rationalization into ODQA models, thus underscoring the importance of passage rationalization.

\section{Related Work}

\paragraph{\textbf{Open Domain Question Answering (ODQA).}}
The prevailing approach to ODQA involves using a retriever to pinpoint relevant passages for a given question from a vast database, such as Wikipedia, and then employing a reader to generate the final answer. This is achieved by integrating the retrieved passages and question with a large pretrained language model. Retrievers commonly use methods ranging from string-matching algorithms like BM25, to dense retrievers such as DPR \citep{DPR}, and semi-parametric retrievers like SEAL \citep{SEAL}. The reader models fall into two primary categories: Extractive readers, such as DPR-reader \citep{DPR}, identify the answer spans from the retrieved passages, while Generative readers, including the Fusion-in-Decoder model (FiD) \citep{FiD} and the Retrieval-Augmented Generation model (RAG) \citep{rag}, generate the answer in a sequence-to-sequence manner.

Our work seeks to enhance the reasoning ability of the FiD reader without modifying the retriever. To this end, KG-FiD \citep{KG-FiD} uses knowledge graphs to rerank and concatenate related passages for improved performance and efficiency. GRAPE \citep{GRAPE} incorporates knowledge graphs into FiD by integrating the Relation-aware graph neural network into the encoder. R2-D2 \citep{R2D2} combines a passage reranker, an extractive reader, and two generative readers into a single comprehensive ensemble model. Unlike these approaches, our work does not involve the use of external knowledge or alternate model architectures, instead focusing on spurious patterns and reader rationale capabilities.

\paragraph{\textbf{Rationale.}}
Recently, spurious patterns have come into the spotlight in NLP \citep{fooling, Jo2017MeasuringTT}, demonstrating a significant impact on model generalization \citep{kaushik2020learning, kaushik2021learning}. Various strategies have been implemented to curb spurious features in tasks like sentiment analysis \citep{lu2022rationale,yang-etal-2021-exploring}, NER \citep{zeng-etal-2020-counterfactual,yang2022factmix}, NLI \citep{DA4NLI} and more \citep{Wang2020RobustnessTS}. 

Our work shares a common goal of overcoming spurious patterns and prioritizing causal features, but it distinguishes itself by using an encoder to identify causal features instead of data augmentation. To the best of our knowledge, we are the first to incorporate rationalization into ODQA.

\citet{asai-etal-2022-evidentiality} also devise multi-task learning methods to train the model to select evidential passages during answer generation, a technique somewhat similar to ours. However, our work differs in two fundamental ways:
1. We strive to guide the decoder with a learnable embedding, which they do not. This approach results in superior performance with an accuracy of 50.7 vs 49.8 on NQ and 69.6 vs 67.8 on TQ.
2. We analyze the rationale ability of our RFiD model, explaining the performance gain and aligning with our motivation, which we consider a significant contribution of this paper.

\paragraph{LLMs in ODQA}
Initial attempts to employ Pretrained Language Models (PLMs) to directly answer open-domain questions without retrieval reported inferior performance compared to DPR+FiD \citep{yu2023generate,wang-etal-2021-generative,rosset2021pretrain}. However, with the advent of Large Language Models (LLMs) like ChatGPT and others, the promise of directly answering open questions based solely on internal parameters became increasingly feasible \citep{shi2023replug}. 

A study by \citet{wang2023evaluating} manually evaluated the performance of LLMs, including ChatGPT-(3.5/4), GPT-3.5, and Bing Chat, alongside DPR+FiD on Natural Questions (NQ) and TriviaQA (TQ) test sets. The results revealed that while FiD surpassed ChatGPT-3.5 and GPT-3.5 on NQ and GPT-3.5 on TQ, the combination of DPR+FiD still showcased considerable potential in the era of LLMs.

\section{Method}
In this section, we explain the baseline Fusion-in-Decoder (FiD) model and our Rational-FiD (RFiD) model. The RFiD model uses a passage-level classifier on top of each FiD encoder to determine whether the corresponding passage is a rationale for the question. It guides the decoder with a rationale embedding concatenated to the encoder hidden states, as well as a multi-task framework to blend FiD training with rationale training.

\subsection{Fusion-in-Decoder for ODQA}
\label{baseline}

The overall input to the reader is a question and $K$ retrieved passages. We feed the text sequence concatenated by the question and one passage to each encoder, and the concatenation detail is in Appendix~\ref{appendix:baseline}.
Formally, for the pair of the question and the $k_{th}$ passage, the input textual sequence $X_{k}$ is  as $x_{k,1}, \ldots, x_{k,i}, \ldots, x_{k,{L}}$, where $x_{k,i}$ represents the $i^{th}$ token and $L$ is the maximum tokens length. We denote the target answer as $Y$, which is also a textual sequence. Therefore, multi-passage QA can be defined as learning the conditional probability $p(Y|X_1,\ldots,X_K;\theta)$, where $\theta$ denotes the model parameters.  Such a model factorizes the conditional probability into $p(y_i|y_1,\ldots,y_{i-1},X_1,\ldots,X_K)$ and is trained in an auto-regressive way. We denote the FiD training loss as $\mathcal{L}_{FiD}$ for further usage.  

The Fusion-in-Decoder (FiD) \citep{FiD} model has been used as a standard baseline for calculating the above probability and find the most probable answer given $K$ question-passage sequences.
FiD has a multi-encoder architecture, with shared parameters. 

\subsection{Passage Rationale}
\label{psg-ratn}

We define a passage as a rational passage to a question if the passage contain at least one answer span from all golden answers, or it is a spurious passage. This is inspired by \citep{DPR} who use a similar method to define positive or negative passages for training the retriever.
We ask the encoders of FiD to distinguish rational and spurious and guide the decoder with the results. 

Formally, we denote $\mathbf{H_{k}} \in \mathbb{R}^{L \times d}$ as the output encoder hidden states of the $k_{th}$ encoder, where $L$ is the maximum tokens length and $d$ is the dimension of hidden states. 
Therefore, the annotation for the input of fusion decoder can be defined as $[\mathbf{H_{1}};\ldots,\mathbf{H_{k}};\ldots,\mathbf{H_K}]$. 
For the $k_{th}$ encoder and its hidden states $\mathbf{H_{k}}$, we apply a binary classifier on the top of the first token's hidden states $\mathbf{H_{k, 1}}$ to distinguish whether the passage is a rationale passage to the question.
The binary classification result of the $k_{th}$ encoder is
\begin{equation}
    \hat{b_k} = Classifier (\mathbf{H_{k, 1}}) \in \mathbb{R}^{2}
\end{equation}

The training loss used for this passage rationale task is can be defined using the Cross Entropy loss
\begin{equation}
    \mathcal{L}_{ratn} = - (b \log(\hat{b}) + (1-b) \log(1-\hat{b}))
\end{equation}
where $b$ is the rational/spurious label while $\hat{b} \in \mathbb{R}^{2} $ is the classification output .

\paragraph{Guiding Decoder.}
After obtaining the output, we guide the decoder with the result by appending additional embeddings to the end of the encoder hidden states and feeding the new encoder hidden states to the decoder.

the prediction label of classification is
$$
pred_k = \arg \max (\hat{b_k}) \in \{0, 1\}
$$

In particular, we use two updatable embeddings
$$
    \mathbf{E^{ratn}_{\{0, 1\}}} \in \mathbb{R}^{2 \times d} 
$$
to represent the passage rationale information, where $d$ is the dimension of encoder hidden states.

The rationale embedding for the $k_{th}$ encoder is 
\begin{equation}
    \mathbf{H_{k,ratn}} =\mathbf{E^{ratn}_{pred_k}}
     \in \mathbb{R}^{d}
\end{equation}

So, the modified encoder hidden states of the $k_{th}$ encoder with rationale embedding is
\begin{equation}
\begin{aligned}
\mathbf{H_k} = [\mathbf{H_{k,1}};\ldots; &\mathbf{H_{k,j}};\dots; \mathbf{H_{k,L}};\mathbf{H_{k,ratn}}] \\
&\in \mathbb{R}^{(L+1) \times d}   
\end{aligned}
 \label{eq:cat_emb}
\end{equation} 
where $L$ is the maximum tokens length and $\mathbf{H_{k,j}}$ is the hidden state of the $j_{th}$ token.


\paragraph{Multi-task Learning.}
We propose a multi-learning framework to train the rationale classifier and the sequence-to-sequence architecture of FiD at the same time.
We define the overall training loss as the sum of the binary classifier training loss $\mathcal{L}_{ratn}$ and the FiD training loss $\mathcal{L}_{FiD}$ :
\begin{equation}
    \mathcal{L}_{total} = \mathcal{L}_{ratn} + \mathcal{L}_{FiD}
\end{equation}

\begin{table}[t]
  \centering
  \small
  \setlength{\tabcolsep}{1.2mm}
  \begin{tabular}{c|c|c|c}
  \toprule
  & train & dev & test \\
  \midrule
  Natural Questions & 79168 & 8757 & 3610 \\
  \midrule
  TriviaQA & 78785 & 8837 & 11313 \\
  \bottomrule
  \end{tabular}
  \caption{Data details of two datasets.
  }
  \label{dataset}
  \vspace{-3mm}
\end{table}

\paragraph{Details.}
We conduct experiments with FiD-base/large and RFiD-base/large on Natural Questions (NQ) \citep{NQ} and TriviaQA (TQ) \citep{TQ}. Their statistics are shown in the Appendix Table~\ref{dataset}.
To avoid the retrieval bias, we follow \citet{FiD, izacard2020distilling} and adopt fixed DPR retrievers to obtaining 100 passages for each question and fix the passages in the following experiments.

\section{Experiment}

\begin{table}[t]
  \centering
  \small
  \setlength{\tabcolsep}{1mm}
  \begin{tabular}{c|c|c|c|c|c}
  \toprule
    &
   \makecell{\#para} &
   \multicolumn{2}{c|}{\makecell[c]{\textbf{NQ}}} & 
   \multicolumn{2}{c}{\makecell[c]{\textbf{TQ}}} \\
  \midrule
  &
  & dev & test
  & dev & test
  \\
  \midrule
  \makecell{RAG} & 626M & - & 44.5 & - & 56.1 \\
  \midrule
  FiD-base & 440M &- & 48.2 & - &  65.0  \\
  FiD-large & 990M & - & 51.4 & - & 67.6 \\
  \midrule
  KG-FiD-base & 443M & - & 49.6 & - & 66.7 \\
  \makecell{KG-FiD-large \\ \citep{KG-FiD}} & 994M & - & 53.4 & - & 69.8 \\
  \midrule
  GRAPE-base  & 454M & - & 48.7 & - & 66.2 \\
  \makecell{GRAPE-large\\ \citep{GRAPE}} & 1.01B & - & 53.5 & - & 69.8 \\
  \midrule
  \multicolumn{6}{c}{Our Implementations } \\
  \midrule
  FiD-base & 440M & 49.3 & 50.2 & 68.6  & 69.0 \\
  \makecell{RFiD-base} & 440M & 50.0 & 50.7 & 69.6 & 69.6 \\
  FiD-large & 990M & 51.6 & 52.8 & 71.6 & 71.9 \\
  \makecell{RFiD-large} & 990M & \textbf{52.5 }& \textbf{54.3} & \textbf{72.7} & \textbf{72.6} \\
  \bottomrule
  \end{tabular}
\caption{
Exact Match scores of different models on the dev/test set of Natural Questions and TriviaQA. The upper part presents results of related reader models, while the lower part presents our implemented FiD and RFiD models. Note that direct comparison between the results of previous work and ours may be affected by differences in retrieved passages.}
\vspace{-2mm}
  \label{results}
\end{table}

\subsection{Main Results}

Table~\ref{results} presents our main results. Our RFiD model outperforms the baseline FiD model on both the Natural Questions (NQ) and TriviaQA (TQ) datasets.

RFiD-large achieved an exact match score of 54.3 on the NQ test set, surpassing the FiD-large baseline score of 52.8 by 1.5 points. This represents a performance increase of roughly 2.8\%. On the TQ test set, RFiD-large scored 72.6, which is 0.7 points higher than the FiD-large score of 71.9, representing an improvement of approximately 0.9\%.
When comparing the base models, RFiD-base scored 50.7 on the NQ test set, which is 0.5 points higher than the FiD-base score of 50.2, corresponding to an approximate improvement of 1.0\%. On the TQ test set, RFiD-base scored 69.6, 0.6 points higher than the FiD-base score of 69.0, reflecting an improvement of around 0.9\%.

These consistent improvements across both base and large models in two different datasets highlight the robustness of our RFiD model in various contexts. Furthermore, these results support the hypothesis that incorporating rationale embeddings in the Fusion-in-Decoder architecture indeed benefits the model's reasoning ability and overall performance.
Additionally, it's worth mentioning that our RFiD model's performance increase is achieved with a negligible increase in parameters. This is demonstrated in the `\#para' column in Table~\ref{results}, further validating the efficiency and practicality of our proposed approach.

In summary, our RFiD model effectively enhances the rationale ability of the Fusion-in-Decoder model, leading to improved performance on open-domain question answering tasks. Our model outperforms both the baseline and other state-of-the-art models on the Natural Questions and TriviaQA datasets, demonstrating the power of our simple yet effective approach.

\begin{table}[t]
  \centering
  \small
  \setlength{\tabcolsep}{1.2mm}
  \begin{tabular}{c|c|c|c|c}
  \toprule
  &
   \multicolumn{2}{c|}{\makecell[c]{\textbf{NQ}}} & 
   \multicolumn{2}{c}{\makecell[c]{\textbf{TQ}}} \\
  \midrule
  & EM & $r_{pos/neg}$ & EM & $r_{pos/neg}$ \\
  \midrule
  FiD-base & 50.3 & 3.71 & 69.0 & 2.14\\
  \midrule
  RFiD-base & 50.7 & 4.31 & 69.6  & 2.32 \\
  \midrule
  FiD-large & 52.8 & 3.82 & 71.9 & 2.17\\
  \midrule
  RFiD-large & 54.3 & \textbf{4.41} & \textbf{72.6} & \textbf{2.52}\\
  \midrule
  \makecell{RFiD-large\\w/o guiding decoder}  & 53.4 & 4.02 & 72.2 & 2.26\\
  \bottomrule
  \end{tabular}
  \caption{The right column shows the ratio of cross-attention scores from the decoder for positive passages versus negative passages. The experiment was conducted on the NQ-test set.
  }
  \label{cross_attn}
  \vspace{-3mm}
\end{table}

\subsection{Cross Attention Analysis}

To evaluate the ability of our RFiD models to distinguish between positive and negative passages, we conducted a cross-attention analysis of the decoder. The principle here is straightforward: better performance would be indicated by more cross-attention focused on positive passages and less cross-attention directed towards negative passages.

The calculations for this analysis are based on a set of equations. To start with, we define the average cross-attention scores on positive passages as follows:
\begin{equation}
    \bar{CA_{pos}} = \frac{1}{N_q} \sum_{q} ( \frac{1}{N^q_{pos}} \sum_{p \in P^q_{pos}} CA_{\{q; p\}})
\end{equation}
where $N_q$ is the number of questions, $P^q_{pos}$ is the set of positive passages on $q$ and $N^q_{pos}$ is its amount, and $CA_{\{q; p\}}$ is the overall cross attention of decoder on the passage $p$ when the question is $q$, which can be calculated as
\begin{equation}
    CA_{\{q; p\}} = \sum_{l}^{N_{ly}} (\sum_{j=1}^{L}  ca_{\{l;j\}}) 
\end{equation}
where $N_{ly}$ is the number of layers, $L$ is the maximum token length and $ca_{\{l; j\}}$ is the cross attention score of the $l_{th}$ layer of decoder on the $j_{th}$ token.

Thus, the ratio $r_{pos/neg}$ of the average cross attention scores of positives passages over the scores of negative passages is 
\begin{equation}
    r_{pos/neg} = \frac{\bar{CA_{pos}}}{\bar{CA_{neg}}}
\end{equation}

The results shown in Table~\ref{cross_attn} reveal that RFiD models have higher $r_{pos/neg}$ values compared to FiD models, suggesting that RFiD focuses more on positive passages and less on negative passages. 
For example, the $r_{pos/neg}$ of RFiD-large is 4.41/2.52 on NQ/TQ, which is 0.39/0.35 higher than the FiD-large. Similarly, for the base model, the improvements are 0.60/0.18 on NQ/TQ. 
The improved ability to identify relevant passages contributes to the overall performance increase seen in our experimental results.

\subsection{Analysis Without Guiding the Decoder}

We also conduct ablation experiment of not guiding the decoder, which means in the Equation~\ref{eq:cat_emb}, the $\mathbf{H_k} = [\mathbf{H_{k,1}};\ldots;\mathbf{H_{k,L}}]$.

The results are displayed in the last row of Table~\ref{cross_attn}.
`RFiD-large w/o guiding decoder' achieves 53.4 and 72.2 EM on the NQ-test and TQ-test, respectively, outperforming the baseline FiD-large by 0.6 and 0.3 EM. The $r_{pos/neg}$ values are 4.02 and 2.26, respectively, which are also higher than the baseline. The results suggest that even without explicitly guiding the decoder, encouraging encoders to discern rationales can still improve performance and the model's ability to identify rationales. This may be because the encoders implicitly encode the rationale information into the hidden states and feed it to the decoder.

\subsection{Case Study} 
As depicted in Figure~\ref{main_example}, the baseline FiD identifies the incorrect answer "March 16, 2002" by referring to spurious passages (shown in blue). It confuses "Disney Art of Animation Resort" with "Art of Disney Animation". In the top 100 retrieved passages, there are many passages about the latter but only one rationale passage (shown in red) about the former. However, our RFiD model can distinguish the only rationale passage from all the spurious ones and mark it for the decoder via an explicit embedding. This enables the decoder to focus more on the rationale, leading to the correct answer.

\section{Conclusion}
In this study, we sought to rationalize the reader component of open question answering by introducing an explicit mechanism to identify potential answer-containing passages among the retrieved candidates. Experimental results show that our RFiD model effectively improves the rationale ability of the Fusion-in-Decoder model, leading to enhanced performance on ODQA tasks. Additionally, it outperforms the baseline and other concurrent models on both the Natural Questions and TriviaQA datasets, with only a minimal increase in the number of parameters.

\section*{Acknowledgement}
We thank for Linyi Yang, Sirui Cheng for their generous help and discussion.
This publication has emanated from research conducted with the financial support of the Pioneer and ``Leading Goose" R\&D Program of Zhejiang under Grant Number 2022SDXHDX0003.

\section*{Limitations}
The method of identifying rational and spurious passages sometimes makes mistakes when 

1. one passage actually contains one golden answer but the content slightly differ the golden answer span, for example, `Messi' vs `Lionel Messi' vs `Lionel Andrés Messi' vs `Lionel Andres Messi'; 

2. one passage actually does not relate the answer but the answer span is too common and appears in the passage, such as `2'.

We just use only seed=0 for the experiments.
\section*{Ethics Statement}
There are no known potential risks.

\bibliography{anthology,acl2023}
\bibliographystyle{acl_natbib}

\appendix

\section{Experiments}

\subsection{Data Process}
\label{appendix:baseline}
Following \citet{FiD}, we concatenate the question and the passage in the form of ``Question : <question> ; Title : <title> ; Context : <context> ''.
\subsection{Implementation Details}
We conduct our experiments on 2 A100-80G-SXM GPUs.

In training, the optimizer is AdamW and the learning rate is 1e-4 with weight decay rate as 0.01; the batch size is 64 and the total training step is 320k.

In evaluating, the eval step is 10k and the best-dev checkpoint is used for the test.

\end{document}